\documentclass[sn-jnl,iicol]{sn-jnl}

\usepackage{graphicx}
\usepackage{amssymb}
\usepackage{stfloats}
\usepackage{amsmath}
\usepackage{url}
\usepackage{multirow}
\usepackage{float}
\usepackage{subfig}
\usepackage{array}
\usepackage{tabularx}
\usepackage{graphicx}
\usepackage{xcolor,array}
\usepackage{comment}
\usepackage{adjustbox} 
\usepackage{makecell} 
\usepackage{arydshln} 
\usepackage[numbers]{natbib}
\usepackage{hyperref}
\usepackage[utf8]{inputenc}
\usepackage{booktabs}%
\usepackage{array}
\usepackage{multirow}
\usepackage{balance}

\DeclareUnicodeCharacter{015F}{\c{s}}
\DeclareUnicodeCharacter{015F}{\c{s}}  

\DeclareUnicodeCharacter{011F}{\u{g}}
\DeclareUnicodeCharacter{0130}{\.{I}} 

\definecolor{backgG}{RGB}{255, 255, 153}
\definecolor{tagtxtG}{RGB}{102, 102, 0}
\definecolor{backgPc}{RGB}{179, 255, 179}
\definecolor{tagtxtPc}{RGB}{0, 102, 0}
\definecolor{backgPw}{RGB}{255, 179, 179}
\definecolor{backgPw}{rgb}{0.0, 1.0, 1.0}
\definecolor{tagtxtPw}{RGB}{0.0, 1.0, 1.0}
\definecolor{backgPo}{rgb}{0.0, 1.0, 1.0}
\definecolor{tagtxtPo}{RGB}{102, 0, 0}
\definecolor{backgPm}{rgb}{0.98, 0.81, 0.69}
\definecolor{tagtxtPm}{RGB}{0,1,1}

\theoremstyle{thmstyleone}%
\theoremstyle{thmstyletwo}%

\theoremstyle{thmstylethree}%

\newcommand{\myorcid}[1]{\href{https://orcid.org/#1}{\includegraphics[width=8pt]{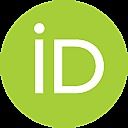}}}

\begin{document}
\title{Automatic Generation of Titles for Research Papers Using Language Models}

\author*[1]{\fnm{Tohida} \sur{Rehman}\myorcid{0000-0002-3578-1316}}\email{tohidarehman.it@jadavpuruniversity.in}

\author[2]{\fnm{Debarshi Kumar} \sur{Sanyal} \myorcid{0000-0001-8723-5002}}\email{debarshi.sanyal@iacs.res.in}

\author[1]{\fnm{Samiran} \sur{Chattopadhyay} \myorcid{0000-0002-8929-9605}}\email{samirancju@gmail.com}

\affil*[1]{Jadavpur University, Kolkata-700106, India}

\affil[2]{Indian Association for the Cultivation of Science, Kolkata-700032, India}


\abstract{
The title of a research paper conveys its primary idea and, occasionally, its conclusions in a clear and concise manner. Choosing an appropriate title is often challenging, and automated title generation can assist authors in this task. In this work, we propose a technique to generate paper titles from abstracts using open-weight pre-trained and large language models. We use the CSPubSum and LREC-COLING-2024 datasets and introduce a new dataset, SpringerSSAT, curated from four Springer journals in the social sciences. Additionally, we use GPT-3.5-turbo in a zero-shot setting to generate titles. Model performance is evaluated with ROUGE, METEOR, MoverScore, BERTScore, and SciBERTScore metrics. Our experiments show that fine-tuned PEGASUS-large outperforms other models, including fine-tuned LLaMA-3-8B and zero-shot GPT-3.5-turbo, across most metrics. We further demonstrate that ChatGPT can generate creative paper titles. Overall, AI-generated titles are generally appropriate and reliable.}

\keywords{Natural language generation, pre-trained language models,  large language models,  evaluation, scholarly publishing} 

\maketitle

\section{Introduction}
\label{intro}
Coming up with a concise and informative title is one of the most important aspects of writing a research paper. In academic publishing, a title should be engaging, enabling readers to quickly grasp the paper's key insights. At first glance, readers should be able to clearly understand the main contributions of the paper. An effective title contains keywords that highlight the main contributions of the paper, and sentences in the body that reuse these terms often reflect the paper's central themes and focus. Numerous studies have examined the relationship between a paper’s title and its readership, as measured by citations and downloads \cite{jamali2011article,letchford2015advantage,rostami2014effect}. Shorter titles have been reported to attract more citations \cite{letchford2015advantage}.

The remarkable success of neural pre-trained language models (PLMs), particularly large language models (LLMs), in various natural language processing (NLP) tasks raises the question of whether they can also generate titles for research papers. Generating a title is challenging because it must succinctly convey the main contribution of a paper while remaining brief and appealing to a broad audience. Consequently, automatic title generation can be viewed as a form of abstractive text summarization, which aims to capture and represent the most salient aspects of a document in a compact form.

In this study, we present a method to generate titles of research papers from their abstracts using PLMs. We fine-tune several PLMs (T5, BART and PEGASUS) on datasets containing abstracts paired with their corresponding titles. We define a PLM with more than a billion parameters as an `LLM'. 
We also evaluate zero-shot title generation using prompts with both pre-trained and fine-tuned LLaMA-3 models, as well as GPT-3.5-turbo. While the PLMs and LLaMA-3 are open-weight models, GPT-3.5-turbo is a closed-weight model. Additionally, we introduce a new dataset, \textbf{SpringerSSAT}, curated from four Springer journals in the social sciences, to evaluate model performance. AI-generated titles can be particularly helpful for non-native English speakers and early-career researchers, who often struggle to construct appropriate titles. These automatically generated titles can subsequently be refined by authors to better align with their individual styles and preferences.

This work extends our previous study \cite{rehman2024can} by introducing the \textbf{SpringerSSAT} dataset, evaluating additional PLMs and LLMs, examining the potential of ChatGPT for generating stylistically creative paper titles, and analyzing variability in human-written titles.

\subsection{Research Objective and Questions}
The primary goal of this research is to investigate how effectively pre-trained and large language models can generate titles of a research paper from abstracts. We will now discuss the specific problems that we aim to address in this work:

\begin{enumerate}
   \item [\textbf{RQ1.}] How accurately can PLMs (such as T5, BART, and  PEGASUS) and LLMs (such as LLaMA-3-8B and GPT-3.5-turbo) generate titles from abstracts?
   \item [\textbf{RQ2.}] Does fine-tuning an LLM on domain-specific data improve the quality of generated titles in terms of coherence, relevance, and factual correctness?
   \item [\textbf{RQ3.}] How well do fine-tuned models work on unseen datasets, such as LREC-COLING-2024, without additional fine-tuning?
   \item [\textbf{RQ4.}] Can ChatGPT generate creative and stylistically diverse titles comparable to human-written titles?
   \item [\textbf{RQ5.}] Are different researchers likely to provide different titles for the same paper? How well do these alternative titles align with both the original author-written titles and those generated by models trained on the author-written titles?
\end{enumerate}
\subsection{Main Contributions}
The main contributions of this paper are as follows:
\begin{enumerate}
    \item We fine-tune the pre-trained language models (PLMs) \textbf{T5-base} \cite{JMLR:v21:20-074}, \textbf{BART-base} \cite{lewis-etal-2020-bart}, and \textbf{PEGASUS-large} \cite{10.5555/3524938.3525989} using the \textbf{CSPubSum} dataset to automatically generate research paper titles from abstracts. We also employ the open-source large language model (LLM) \textbf{LLaMA-3-8B} \cite{touvron2023llama, llama3modelcard} and evaluate its generation quality under two settings -- fine-tuned and non-fine-tuned. In addition, \textbf{GPT-3.5-turbo} is used in a zero-shot configuration to perform the same task. 
    \item We introduce a new dataset of 3,473 abstract–title pairs, named \textbf{SpringerSSAT}, curated from Springer conference and journal papers. All the above PLMs and the LLM \textbf{LLaMA-3-8B} are fine-tuned on this dataset to evaluate their capability for automatic title generation.

    \item We also utilize our previously released dataset, \textbf{LREC-COLING-2024} \cite{rehman2024can}, comprising 1,000 abstract–title pairs, to assess model generalization. Specifically, we evaluate models fine-tuned on both \textbf{CSPubSum} and \textbf{SpringerSSAT} datasets on this corpus without further training.
    
    \item Model performance is evaluated using standard automatic metrics -- \textbf{ROUGE} \cite{lin2004rouge}, \textbf{METEOR} \cite{banerjee2005meteor}, \textbf{MoverScore} \cite{zhao-etal-2019-moverscore}, and \textbf{BERTScore} \cite{zhang2019bertscore} as well as its domain-specific variant \textbf{SciBERTScore}. We further assess factual accuracy at the entity level using the precision-source and F1-target metrics proposed by Nan et al.\ \cite{nan-etal-2021-entity}. Additionally, a manual evaluation is conducted by human annotators on a subset of generated titles.
    
    \item Our experiments demonstrate that \textbf{PEGASUS-large} achieves superior performance across most metrics while maintaining significantly fewer parameters than larger LLMs such as GPT-3.5-turbo and LLaMA-3-8B.
    
    \item We further explore whether ChatGPT can generate creative and stylistically diverse titles for research papers. The generated titles are compared against author-written titles and evaluated by human annotators. 
    
  \item All fine-tuned models and the previously released dataset, \textbf{LREC-COLING-2024}, are publicly available on our Hugging Face account\footnote{\url{https://huggingface.co/datasets/TRnlp/LREC-COLING-2024-Abstract-Title}}. 
  We have publicly released the  \textbf{SpringerSSAT}  dataset on Hugging Face\footnote{\url{https://huggingface.co/datasets/TRnlp/SpringerSSAT}}. Supplementary materials supporting this study are available on GitHub\footnote{\url{https://github.com/tohidarehman/Title-Generation-ResearchPapers}}.
\end{enumerate}

We briefly outline the structure of the paper. Section~\ref{Literature Survey} explains prior work on research paper title generation and related NLP tasks. Section~\ref{Datasets} describes the datasets used in this research, including CSPubSum, LREC-COLING-2024, and the newly contributed SpringerSSAT dataset, providing dataset statistics as well as the train, validation, and test splits. In Section~\ref{Methodology}, we present brief overviews of the models we utilized for automatic title generation from paper abstracts. Section~\ref{Experimental Setup} outlines the data preprocessing steps, implementation details, and the evaluation metrics used in this research. Section~\ref{results} reports quantitative results, comparing the performance of fine-tuned models on CSPubSum and SpringerSSAT (\textbf{addressing RQ1 and RQ2}) and examining cross-domain generalization to LREC-COLING-2024 (\textbf{RQ3}). Section~\ref{sec:caseStudies} provides examples of titles generated by the different models and a manual evaluation of a subset of generated titles. Section~\ref{sec:CreativeTitles} explores the generation of creative titles using ChatGPT (\textbf{RQ4}). Section~\ref{ExpertEvaluation} analyzes the scenario where the same abstract is titled by multiple domain experts and whether they align with the outputs of the trained models  (\textbf{RQ5}). Section~\ref{Limitations} discusses the limitations of our research, and Section~\ref{conclusion} concludes the paper and outlines directions for future research. 

\section{Literature Survey}
~\label{Literature Survey}
Research on automatic text summarization has been ongoing for decades. One of the first approaches was proposed by Luhn et al. ~\cite{luhn1958automatic}, which focused on extractive summarization by selecting sentences with recurring important terms, not important sentences with generic words. Luhn's method was used for summarizing technical papers and magazine articles. Years later, Lloret et al. ~\cite{lloret2013compendium} proposed the COMPENDIUM system to generate biomedical abstracts and combined extractive, and hybrid extractive-abstractive methods to assess the summarization methods and their effectiveness.

The field of abstractive summarization achieved notable improvements with sequence-to-sequence models~\cite{Sutskever-2014-sequence}, attention-based encoders with beam search and RNNs~\cite{bahdanau2015neural, nallapati2016abstractive}, and pointer-generator networks with coverage mechanisms~\cite{See2017GetTT}, which were mostly applied to news articles. Later, the transformer architecture~\cite{vaswani2017attention} changed the landscape of NLP research, leading to pre-trained models like BART~\cite{devlin2018bert}, T5~\cite{raffel2019exploring}, and PEGASUS~\cite{zhang2020pegasus}. These models are trained on large and general-purpose text corpora in a self-supervised way, which helps them learn linguistic patterns and knowledge, and then they can be fine-tuned for specific tasks in different domains.

In the context of title generation, several studies have focused on generating titles for news articles from their content. Tan et al.~\cite{tan2017neural} proposed a coarse-to-fine approach, where important sentences are identified using hierarchical attention within an encoder-decoder framework. Their model was evaluated on the New York Times (NYT) and DUC-2004 datasets. Putra et al.~\cite{putra2017automatic} proposed a model that used sentence-level rhetorical categories to identify research purpose and method, tested on abstracts from computational linguistics and chemistry papers. 

For academic papers, Mishra et al.~\cite{mishra2021automatic} proposed a method that first generates candidate titles using a pre-trained model, then selects the most suitable title and refines it to ensure semantic and syntactic accuracy. Their experiments used datasets from arXiv~\footnote{\url{https://tinyurl.com/y9pu6xyp}}, ACL~\cite{wang2018paper}, and ICMLA~\cite{vallejo2019dataset}. Similarly, Bikku et al.~\cite{bikku2023generating} introduced a GPT-3 based framework for research paper title generation, which selects titles from a candidate pool and applies filtration modules. They evaluated model performance on the arXiv dataset using BLEU, ROUGE, and human judgment.

More recently, Liu et al. \cite{liu2022oag} pre-trained a transformer encoder, OAG-BERT, on computer science research  papers (including its metadata) and used it to produce paper titles, which were deemed acceptable by researchers. However, comprehensive evaluation of these generated titles using automatic metrics has not been done. Another emerging area of interest is the generation of research highlights from paper abstracts \cite{rehman2021automatic, rehman-etal-2022-named, rehman2023research, 10172215}. Research highlights are bullet-point summaries that accompany the abstract and emphasize the primary findings of the paper. 
In contrast to this line of work, we focus on generation of titles for research papers across multiple domains and employ a diverse set of encoder-decoder and decoder-only transformer models for this task. We also perform a detailed evaluation of the generated titles. Our work aligns with contemporary research that explores the potential of GPT and other artificial intelligence tools in education, research, and scholarly publishing \cite{lin2023and,ciaccio2023use}. 

Finally, recent datasets have further supported research in this area. B\"{o}l\"{u}c\"{u} et al.~\cite{bolucu2025modest} introduced MoDeST, a multi-domain and multilingual dataset for scientific title generation, noting that abstracts serve as the most effective input for this task. In this study, we introduce the SpringerSSAT dataset, curated from multiple Springer journals in the social sciences, to evaluate the performance of both PLMs and LLMs on the title generation task. Our previous work~\cite{rehman2024can} presented the LREC-COLING-2024 dataset and focused on automatic title generation on this dataset as well as CSPubSum. The present journal version extends that study with additional models, datasets, and evaluation methodologies.

\section{Datasets}~\label{Datasets}
In this section, we describe the curation process and key characteristics of the three datasets: CSPubSum, LREC-COLING-2024, and SpringerSSAT.

\subsection{Curation}
We use the CSPubSum dataset provided by Collins et al.~\cite{collins2017supervised}, which includes URLs for 10,147 computer science papers from ScienceDirect\footnote{\url{https://www.sciencedirect.com/}}. 
Given this dataset, we crawled and organized the data such that each example forms a pair comprising an \textit{abstract} and its corresponding \textit{author-written title}. The dataset is divided into 8,120 papers for training, 1,014 for validation, and 1,013 for testing.

We additionally crawled 1000 accepted papers from LREC-COLING 2024 that are available on ACL Anthology\footnote{\url{https://aclanthology.org/events/coling-2024/}} under the CC BY 4.0 license \cite{rehman2024can}. This dataset, hereafter referred to as LREC-COLING-2024, consists of (\textit{abstract}, \textit{author-written title}) pairs for each paper. 

We have constructed a new dataset with papers from four Springer journals in the domain of social sciences. Specifically, we collected 1,152 abstract-title pairs from \textit{SN Social Sciences}, 1,135 pairs from \textit{Society}, 416 pairs from \textit{Race and Social Problems}, and 770 pairs from \textit{Social Justice Research}. The dataset consists of 3,473 abstract–title pairs, which we refer to as the SpringerSSAT dataset. We split the dataset into 2,778 papers for training, 347 for validation, and 348 for testing.
For \textit{SN Social Sciences}, we collected all research papers from Volumes~1--4 (Issues~1--12) and Volume~5 (Issues~1--10). For \textit{Society}, all research papers from Volumes~44--61 (Issues~1--6) and from Volume~62 (Issues~1--4) were selected. For Volumes~1--43, only abstracts and titles of papers available through online open access were included. For \textit{Race and Social Problems}, we collected research papers from Volumes~1--16 (Issues~1--4) and Volume~17 (Issues~1--3). For \textit{Social Justice Research}, we included all four issues from each of Volumes~1--37, and Issues~1--3 from Volume~38. Overall, the dataset includes all openly available abstract-title pairs published up to September 2025 in these four journals. These four Springer journals were selected due to their broad thematic coverage within the social sciences and the availability of structured abstract--title metadata. Collectively, they focus on sub-fields such as sociology, race and ethnicity studies, social policy, and social justice research. However, as these journals are published by a major international academic publisher, the dataset may reflect broader publication patterns typical of internationally indexed social science research. Consequently, certain regions or research traditions may be more prominently represented than others. Additionally, since some journals were established more recently (e.g., \textit{SN Social Sciences}), the temporal distribution of articles is skewed toward the past decade. These factors should be considered when interpreting model generalization and cross-domain performance.

\subsection{Data Characteristics}

The CSPubSum and LREC-COLING-2024 datasets both comprise computer science papers. However, the latter consists exclusively of NLP papers while the former is more diverse. On the other hand, SpringerSSAT contains papers from the social science.

Table~\ref{Table:token_stats} shows the average and maximum token counts for abstracts and titles across all datasets. Token counts are computed using a whitespace-based tokenization approach. The average number of tokens in a title is 12 for both CSPubSum and SpringerSSAT and 11 for LREC-COLING-2024, although some titles are longer and a few very short. A few titles even consist of multiple sentences; for example, one title in CSPubSum reads: ``Comparative statics effects independent of the utility function. When do we act the same way under risk?''. However, approximately 82\% of papers in CSPubSum and 90\% of papers in LREC-COLING-2024 have titles with no more than 15 tokens. In SpringerSSAT, 78.34\% of papers have title length within 15 tokens. 
Some titles in SpringerSSAT are particularly long, such as, ``Comparing household heads’ perception of climate change variability with meteorological trends and understanding mitigation measures to combat the adverse effects in coastal areas of Bangladesh'', ``An exploration of how the disruption of mainstream schooling during the COVID-19 crisis provided opportunities that we can learn from so that we may improve our future relationship with the more-than-human world'', and ``What Matters More, Maternal Characteristics or Differential Returns for Having Them? Using Decomposition Analysis to Explain Black-White Racial Disparities in Infant Mortality in the United States''. Note that while CSPubSum and SpringerSSAT have train, val, and test splits, LREC-COLING-2024 is used only as a test set.

For automatic title generation, the \textit{author-written titles} are treated as the \textit{gold standard} references in all cases. In one set of experiments, we fine-tuned the selected language models on the training subset of CSPubSum, but evaluated them on the test subsets of both CSPubSum and LREC-COLING-2024. While CSPubSum covers a large range of topics in computer science, most papers belong to the pre-transformer era~\cite{vaswani2017attention}, since the impact of transformers began to emerge after 2017, the year CSPubSum was released. We aim to assess how well the trained models perform on a different and more recent dataset that they were not explicitly trained on; this motivated the curation of LREC-COLING-2024. 

We also fine-tuned the selected language models on the training subset of the SpringerSSAT dataset and evaluated their performance on the SpringerSSAT test set as well as the LREC-COLING-2024 corpus. Since these two corpora focus on very different domains, these experiments provide insights into the cross-domain generalization capabilities of the models.

\begin{table*}[!htbp]
\centering
\caption{Token count statistics for abstracts and titles in the three datasets. Note that the LREC-COLING-2024 corpus is used only as a test set.}
\label{Table:token_stats}
\begin{tabular}{lcccc}
\toprule
Dataset & \multicolumn{2}{c}{Abstract (Train; Val; Test)} & \multicolumn{2}{c}{Title (Train; Val; Test)} \\
\cmidrule(lr){2-3} \cmidrule(lr){4-5}
 & Avg & Max & Avg & Max \\
\midrule
CSPubSum & 185; 180; 194 & 994; 1304; 1166 & 12; 14; 12 & 60; 32; 26 \\
SpringerSSAT & 165; 163; 173 & 1359; 1257; 995 & 12; 12; 13 & 32; 27; 27 \\
LREC-COLING-2024 & 0; 0; 162 & 0; 0; 290 & 0; 0; 11 & 0; 0; 25 \\

\bottomrule
\end{tabular}
\end{table*}

\section{Methodology}~\label{Methodology}
\begin{figure*} [!htbp] 
\centering
\includegraphics[width=0.9\linewidth,height=10cm]{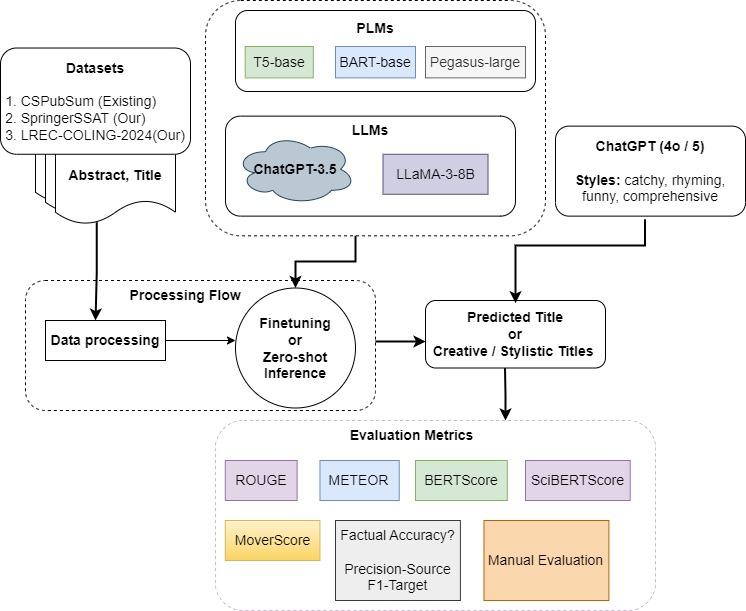}
\caption{End-to-end experimental framework for automatic title generation, illustrating model fine-tuning, creative generation, and multi-faceted evaluation methodology.}
\label{fig:model_diagram}
\end{figure*}

In this section, we describe the various language models that we use for automatic title generation. In particular, we have utilized the following pre-trained models:
\begin{enumerate}
    \item \textbf{T5-base} \cite{JMLR:v21:20-074}: An encoder-decoder model based on the original transformer architecture~\cite{vaswani2017attention}, T5 formulates every NLP  task, such as translation, question answering, and  classification, as a ``text-to-text'' problem, so a single architecture can handle them all in a unified and consistent manner. To pre-train the model, random text spans are masked, and the model is trained to generate them. T5-base contains approximately 220M parameters. 
    \item \textbf{BART-base} \cite{lewis-etal-2020-bart}: A denoising autoencoder, it combines bidirectional and auto-regressive transformers exemplified by BERT \cite{devlin2018bert} and GPT \cite{brown2020language}, respectively. To pre-train BART, the input text is first corrupted with a noising function, and then the model is optimized to reconstruct the original text. It is particularly useful for text generation problems. {BART-base} consists of 139M parameters. 
    
    \item \textbf{PEGASUS-large} \cite{10.5555/3524938.3525989}: It is a transformer-based encoder-decoder model that is pre-trained on large text collections with an objective function specifically focused on summarization. In particular, pre-training the model involves masking \textit{important} sentences from an input document and generating them as one output sequence. PEGASUS-large contains 568M parameters. 
    
    \item \textbf{LLaMA-3-8B}: We have employed the pre-trained {LLaMA-3-8B} model\footnote{\url{https://ai.meta.com/blog/meta-llama-3/}}, which consists of 8 billion parameters. We have used it in two ways: with fine-tuning and without fine-tuning. The LLaMA series of models \cite{touvron2023llama} are decoder-only, transformer-based large language models pre-trained exclusively on publicly available datasets (in contrast to the GPT series). 

    \item \textbf{GPT-3.5-turbo}: We have used the \texttt{gpt-3.5-turbo-1106} model via the OpenAI API\footnote{\url{https://platform.openai.com/docs/models/gpt-3-5-turbo}}. Its web version is commonly known as ChatGPT-3.5\footnote{\url{https://chatgpt.com/}}. It is a decoder-only model pre-trained on massive text corpora and further fine-tuned using instruction tuning and reinforcement learning from human feedback. It is a successor to GPT-3, a large language model with 175B parameters. We have used a prompt-based zero-shot in-context learning setup, where we prompt the model to generate a title given the abstract.
\end{enumerate}
We have fine-tuned the first four models on the train subset of CSPubSum and SpringerSSAT datasets separately. Note that although the above models come in multiple sizes, we have chosen the smallest versions that were freely available at the time of this research,  due to constraints on our computational resources. The paper that introduced the PEGASUS model \cite{10.5555/3524938.3525989} discusses both a base and a large model, but we could only find the checkpoints for PEGASUS-large on Hugging Face  and thus, we selected this version. Although GPT-4 is more powerful than GPT-3.5-turbo, the former is only available to paid subscribers, and therefore, we do not use it for this research. However, we anticipate that, as with other deep learning applications, using larger models will improve the quality of the generated titles. 
Figure \ref{fig:model_diagram} outlines the experimental framework for title generation and evaluation.

\section{Experimental Setup}~\label{Experimental Setup}

\subsection{Data Preprocessing}
We removed extra spaces from the documents in all datasets and retained only the examples in which the abstract length is at least 20 tokens and the title length is at least 3 tokens. Since we imposed a limit of 20 tokens on the generated title length, we required the abstract to be longer than 20 tokens so that the task can be treated as a text summarization problem.

\subsection{Implementation Details}
\label{sec:Impl}

For the CSPubSum dataset, we used three pre-trained models from Hugging Face: {T5-base}\footnote{\url{https://huggingface.co/t5-base}}, {BART-base}\footnote{\url{https://huggingface.co/facebook/bart-base}}, and {PEGASUS-large}\footnote{\url{https://huggingface.co/google/pegasus-large}}. Each of these models was fine-tuned for 5 epochs. We used a batch size of 32 and a learning rate of 4e-5. When fine-tuning or performing inference with the pre-trained models, we set the maximum title length to 20 tokens, which aligns with prior findings that shorter titles are generally more appealing to readers~\cite{letchford2015advantage}.


To generate titles using GPT-3.5-turbo and LLaMA-3-8B, we used the following prompt:

\texttt{Create a concise title from this abstract using at most 20 tokens, highlighting the main contributions and focus. <ABSTRACT>}

Both models received this prompt directly during evaluation, without any in-context examples. Unlike GPT-3.5-turbo, LLaMA-3-8B can be fine-tuned. However, full fine-tuning is extremely resource-intensive. To make things feasible, we applied Low-Rank Adaptation (LoRA), which enables parameter-efficient fine-tuning. In particular, we used the {LLaMA-3-8B}\footnote{\url{https://huggingface.co/unsloth/llama-3-8b-bnb-4bit}}
model in 4-bit precision to save memory. Fine-tuning was carried out for 5 epochs with a training batch size of 32 and an evaluation batch size of 1. The LoRA setup used a rank of 16 ({\tt r=16}) and an identical scaling factor ({\tt lora\_alpha=16}), loaded via {\tt peft\_config}. The same prompt was used throughout training and evaluation.
For comparison, we also used the LLaMA-3-8B without fine-tuning, which we denote as {LLaMA-3-8B$^\ast$}. For GPT-3.5-turbo, we set a temperature of 0.3 when generating titles.

For all models, we limited to 512 input tokens and 20 output tokens (for title generation).

Memory and compute usage were tracked using WandB\footnote{\url{https://wandb.ai/site}}.  Among the smaller models, PEGASUS-large finished in about 23 minutes, BART-base in roughly 7 minutes, and T5-base around 13 minutes. Fine-tuning LLaMA-3-8B on CSPubSum took just under 2 hours. All of this ran on a Tesla A100-SXM4-40GB GPU via \texttt{Colab Pro+}.

For the SpringerSSAT dataset, we separately fine-tuned the same models, T5-base, BART-base, and PEGASUS-large, using the same settings, except the batch size was lowered to 8. Additionally, LLaMA-3-8B was fine-tuned on this dataset; for this training, we used a batch size of 1 and limited the number of epochs to 3 due to resource constraints. 

For SpringerSSAT, the prompt used for LLaMA-3-8B was:

\texttt{Create a concise title from this abstract using at most 20 tokens, highlighting the main contributions and focus. <ABSTRACT>}

We have also included results for the non-fine-tuned LLaMA-3-8B variant ({LLaMA-3-8B$^\ast$}) for comparison. Fine-tuning times for the smaller models were approximately as follows: T5-base: 23 minutes, BART-base: 13 minutes, PEGASUS-large: 50 minutes (all for 5 epochs). Fine-tuning LLaMA-3-8B on SpringerSSAT  dataset took around 4 hours and 20 minutes for 3 epochs using a T4 GPU via the \texttt{Colab} free version.

\subsection{Evaluation Metrics}
We use the commonly used automatic text summarization evaluation metrics, including ROUGE \cite{lin2004rouge}, METEOR \cite{banerjee2005meteor}, MoverScore \cite{zhao-etal-2019-moverscore}, BERTScore \cite{zhang2019bertscore} and SciBERTScore \cite{beltagy2019scibert}, to assess the quality of the generated titles with reference to the author-crafted  titles.  ROUGE scores measure $n$-gram overlap between the generated titles and the ground-truth titles. We have used ROUGE-1, ROUGE-2, and ROUGE-L where the first uses unigram overlap, the second bigram overlap, and the last compares the longest common subsequence between the generated title and the golden title. METEOR measures sentence-level accuracy based on the alignment between the generated text and the reference text. In contrast, MoverScore, BERTScore, and SciBERTScore aim to measure the \textit{semantic similarity} between the output and the true title, using latent representations (embeddings) of the texts. MoverScore combines  Word Mover's Distance \cite{kusner2015word} and contextualized embeddings of the output and the ground-truth titles for semantic matching \cite{zhao-etal-2019-moverscore}. BERTScore calculates the cosine similarity between the BERT embeddings of the words in the two text sequences \cite{zhang2019bertscore}. Since the current application is in the domain of computer science, we have developed a variant of BERTScore, where we use SciBERT \cite{beltagy2019scibert} to generate the embeddings, then compute their cosine similarity as in BERTScore; we call the modified metric SciBERTScore.

However, these metrics are inadequate to quantify \textit{factual consistency} \cite{kryscinski-etal-2020-evaluating}, which, in our context, refers to whether the entities present in the generated titles also occur in the corresponding abstracts and the extent to which these entities overlap with those in the author-written titles. Therefore, we employ three new metrics, introduced by \cite{nan-etal-2021-entity}, to assess the factual consistency of the generated titles. 

Let us call the author-written title as the target $t$, the model-generated title as the hypothesis $h$ and the input abstract as the source $s$. 
We define $\mathcal{N}(t)$ as the count of named entities in the target (author-written title) and $\mathcal{N}(h)$ as the count of named entities in the hypothesis (model-generated title). Then, $\mathcal{N}(h \cap s)$ is the number of entity matches between the generated title and the input source (abstract). Some named entities in the title span multiple words. We used the scispaCy model \texttt{en\_core\_sci\_sm}\footnote{\url{https://allenai.github.io/scispacy/}} to identify entities.
\textbf{Precision-source}, defined as $prec_s= \mathcal{N}(h \cap s) / \mathcal{N}(h) $, 
is a metric used to assess the intensity of hallucination in relation to the \textit{input source (abstract)}. Note that $prec_s$ represents the percentage of entities mentioned in the generated title that can be retrieved from the input source document (abstract).  Low $prec_s$ indicates the possibility of  hallucination in the generated title. However, $prec_s$ does not capture the generated title's entity-level correctness in relation to the \textit{author-written title}. 
Therefore, recollecting that the author-written title is the target $t$, we define target-level entity accuracy in terms of \textbf{precision-target}  
($prec_t)= \mathcal{N}(h \cap t) / \mathcal{N}(h);$    
 \textbf{recall-target}  
($recall_t) = \mathcal{N}(h \cap t) / \mathcal{N}(t);$ and \\ \textbf{F1-target}   
($F1_{t}) = \frac{2*(recall_t*prec_t)}{recall_t+prec_t}.$ 
Here, $\mathcal{N}(h \cap t)$ represents the number of matched named entities in the generated title and the author-written title. 
Note that we can calculate the above mentioned quantities in two ways.
The first considers entity mentions in each document (which may be the source $s$ or target $t$ or hypothesis $h$) as a set so that multiple occurrences of an entity in a document are equivalent to a single occurrence. The other treats  the entity mentions in the input as a multiset, where repeated occurrences are counted separately. In this case, if a metric is defined as $\mu = \mathcal{N}(x \cap y) / \mathcal{N}(x)$, then $\mathcal{N}(x \cap y)$ is computed as follows: for each entity mention in $x$, we check if it occurs in $y$; if it does, we increment the intersection count  $\mathcal{N}(x \cap y)$ (which is initialized to zero) by unity; this approach is followed in \cite{nan-etal-2021-entity}. In the first approach, we denote the metrics as $prec_s^{U}$, $prec_t^{U}$, $recall_t^{U}$, and $F1_t^{U}$, where $U$ indicates that only unique entity mentions are considered. In the second, we represent them as $prec_s^{NU}$, $prec_t^{NU}$, $recall_t^{NU}$, and $F1_t^{NU}$, where $NU$ denotes that repeated mentions are counted separately.
\balance
\section{Results}~\label{results}
\subsection{Comparison of Fine-Tuned Models on CSPubSum}
\label{sec:quanResults}
In this sub-section, we report the results of experiments on the test subset of the {CSPubSum} corpus after the models (except GPT-3.5) were fine-tuned on the train split of CSPubSum. Table~\ref{Table:par_all_types_rouge_meteor_bert-cs} shows the performance in terms of ROUGE, METEOR, and semantic metrics such as MoverScore, BERTScore, and SciBERTScore. Evaluation results using entity-level factual consistency metrics—precision-source ($prec_s^{NU}$, $prec_s^{U}$), precision-target ($prec_t^{NU}$, $prec_t^{U}$), recall-target ($recall_t^{NU}$, $recall_t^{U}$), and F1-target ($F1_t^{NU}$, $F1_t^{U}$) are shown in Table~\ref{Table:entity-scoreCSPuBSum}.

The overall finding is that the \textbf{PEGASUS-large} model achieves the highest scores for all the above metrics except the precision-target metric. It is also observed that for each evaluated metric, the performance differences among the three PLMs -- T5-base, BART-base and PEGASUS-large -- remain relatively small.

\begin{table*}[!h]
\centering
\caption{\small Evaluation of the quality of titles generated from abstracts in \textbf{CSPubSum} test set. Fine-tuning was done on CSPubSum training set. All scores are reported as F1-scores in \%.}
\label{Table:par_all_types_rouge_meteor_bert-cs}
\begin{adjustbox}{width=1.0\linewidth}
{\begin{tabular}{|lccccccc|} \hline
Model Name &ROUGE-1 &ROUGE-2 &ROUGE-L &METEOR  &MoverScore &BERTScore &SciBERTScore\\\hline
T5-base &44.25 &25.04 &38.92 &38.36 &38.09 &89.9 &76.06\\ \hline
BART-base &45.7 &25.97 &40.11 &39.37 &39.75  &90.21 &76.89\\ \hline
PEGASUS-large &\bf{46.75} &\bf{27.13} &\bf{40.67} &\bf{42.61} &\bf{40.43}  &\bf{90.35} &\bf{76.93}\\ \hline
LLaMA-3-8B$^\ast$ &28.4 &12.58 &24.6 &27.17 &21.42  &86.34 &66.65\\ \hline
LLaMA-3-8B &40.8 &21.23 &36.57 &34.5 &37.02  &89.99 &76.41\\ \hline
GPT-3.5-turbo &42.81 &21.16 &36.55 &35.12 &37.39 &88.66 &76.28 \\ \hline
\end{tabular} }
\end{adjustbox}
\end{table*}

\begin{table*}[!h]
\centering
\caption{\small Evaluation of factual consistency of titles generated from abstracts in \textbf{CSPubSum} test set. Fine-tuning was done on CSPubSum training set. All scores are in \%.}
\label{Table:entity-scoreCSPuBSum}
\begin{adjustbox}{width=1.0\linewidth} 
{\tiny
{\begin{tabular}{|ccccccccc|} \hline

Model Name &$prec_s^{NU}$ & $prec_s^{U}$& $prec_t^{NU}$& $recall_t^{NU}$ & $F1_t^{NU}$ & $prec_t^{U}$& $recall_t^{U}$& $F1_t^{U}$ \\ \hline
T5-base &97.1 &97.08 &59.3 &51.82 &52.38 &59.08&51.58 &52.17\\ \hline
BART-base &97.44 &97.39 &\bf{61.52} &52.84 &53.91 &\bf{61.35} &52.56 &53.72  \\ \hline
PEGASUS-large &\bf{98.13} &\bf{98.08} &60.39 &\bf{56.49} &\bf{55.39} &60.17 &\bf{56.21} &\bf{55.18} \\ \hline
LLaMA-3-8B$^\ast$ &67.7 &67.7 &39.19 &40.48 &36.81 &38.96 &39.98 &36.48 \\ \hline
LLaMA-3-8B &78.95 &78.93 &57.54 &46.99 &49.12 &57.44&46.97 &49.08 \\ \hline
GPT-3.5-turbo &92.73 &92.73 &59.79 &50.95 &51.7 &59.79 &51.01 &51.74 \\ \hline
\end{tabular} 
}
}
\end{adjustbox}
\end{table*}

\subsection{Comparison of Fine-Tuned Models on SpringerSSAT}
This subsection presents the performance of the different fine-tuned models on the test subset of SpringerSSAT, after being trained on its corresponding training split. Table \ref{Table:par_all_types_rouge_meteor_bert-SpringerSSAT} shows the performance scores. Entity-level factual consistency evaluation results appear in Table \ref{Table:entity-SpringerSSAT}. We observe that the \textbf{PEGASUS-large} model fine-tuned on the SpringerSSAT dataset achieves the highest scores for all the above metrics, except the precision-source metric, on which T5-base performs better. Similar to the results on CSPubSum, the three pre-trained models -- T5-base, BART-base and PEGASUS-large -- show similar performance across all metrics.

\begin{table*}[!h]
\centering
\caption{\small Evaluation of the quality of titles generated from abstracts in \textbf{SpringerSSAT} test set. Fine-tuning was done on SpringerSSAT training set. All scores are reported as F1-scores in \%.}
\label{Table:par_all_types_rouge_meteor_bert-SpringerSSAT}
\begin{adjustbox}{width=1.0\linewidth}
{\begin{tabular}{|lccccccc|} \hline
Model Name &ROUGE-1 &ROUGE-2 &ROUGE-L &METEOR  &MoverScore &BERTScore &SciBERTScore\\\hline
T5-base &40.34 &19.92 &33.7 &34.49 &28.01 &88.43 &71.6\\ \hline
BART-base &39.1 &18.38 &32.57 &31.88 &27.95 &88.48 &71.8\\ \hline
PEGASUS-large &\bf{42.24} &\bf{20.52} &\bf{35.32} &\bf{34.87} &\bf{29.97} &\bf{88.71} &\bf{72.43}\\ \hline
LLaMA-3-8B$^\ast$ &23.3 &8.61 &18.83 &24.92 &8.76 &85.53 &61.05\\ \hline
LLaMA-3-8B &35.22 &15.96 &30.31 &29.07 &25.96 &88.23 &71.19\\ \hline
\end{tabular} }
\end{adjustbox}
\end{table*}

\begin{table*}[!h]
\centering
\caption{\small Evaluation of factual consistency of titles generated from abstracts in \textbf{SpringerSSAT} test set. Fine-tuning was done on SpringerSSAT training set. All scores are in \%.}
\label{Table:entity-SpringerSSAT}
    \begin{adjustbox}{width=1.0\linewidth} 
{\tiny
{\begin{tabular}{|ccccccccc|} \hline

Model Name &$prec_s^{NU}$ & $prec_s^{U}$& $prec_t^{NU}$& $recall_t^{NU}$ & $F1_t^{NU}$ & $prec_t^{U}$& $recall_t^{U}$& $F1_t^{U}$ \\ \hline
T5-base &\bf{96.7} &\bf{96.68} &48.76 &50.04 &46.38 &48.5  &49.13 &45.91 \\ \hline
BART-base &92.98 &92.93 &49.55 &45.06 &44.82 &49.32 &44.56 &44.55  \\ \hline
PEGASUS-large &95.23 &95.22 &\bf{52.36} &\bf{50.74} &\bf{48.92} &\bf{52.16} &\bf{50.32} &\bf{48.67}  \\ \hline
LLaMA-3-8B$^\ast$ & 57.84 &57.08 &28.93 &49.42 &32.66 &27.95 &45.58 &31.11  \\ \hline
LLaMA-3-8B &81.03 &80.87 &51.95 &41.94 &43.85 &51.68 &41.96 &43.77  \\ \hline
\end{tabular} 
}
}
\end{adjustbox}
\end{table*}

\subsection{Cross-Domain Generalization to LREC-COLING-2024} 
We evaluated the performance of all fine-tuned models on the LREC-COLING-2024 dataset under two fine-tuning scenarios: Case 1 and Case 2. \\ 
\textbf{Case 1:} Models fine-tuned only on the CSPubSum training set. \\
\textbf{Case 2:} Models fine-tuned only on the SpringerSSAT training set.  

As shown in Tables \ref{Table:par_all_types_rouge_meteor_bert-coling} and \ref{Table:entity-scorecoling}, the models fine-tuned on the CSPubSum dataset generally outperform those fine-tuned on the SpringerSSAT dataset, with the exception of the fine-tuned LLaMA-3-8B model, which achieves marginally higher scores on some metrics when fine-tuned on SpringerSSAT dataset. The superior results obtained through CSPubSum fine-tuning indicate that similarity between training and testing domains significantly influences model performance: both CSPubSum and LREC-COLING-2024  datasets contain computer science abstracts, whereas SpringerSSAT consists of social science abstracts. LLaMA-3-8B likely acts as an outlier due to its large-scale pre-training, which provides a broader knowledge base and enables better generalization across domains even without fine-tuning, compared to smaller task-specific models that depend heavily on fine-tuning.

Among all models, {\bf PEGASUS-large} fine-tuned on the CSPubSum dataset achieves the highest scores across nearly all metrics, except for BERTScore. This indicates that its strong performance within the computer science domain transfers effectively to LREC-COLING-2024, despite the model not being explicitly fine-tuned on this dataset. 

It is important to note that two models -- unfine-tuned {LLaMA-3-8B$^\ast$} and {GPT-3.5-turbo} -- do not have Case 1 or Case 2 evaluations, as they were not fine-tuned on either CSPubSum or SpringerSSAT. Their reported results reflect performance without any domain-specific dataset dependency.

\subsection{Overall Analysis}

The results in the preceding subsections indicate that the \textbf{PEGASUS-large} model achieves the best performance on most metrics across all datasets. Specifically, it generates titles that are very close to the author-written ones. 
We attribute the superior performance of PEGASUS-large to its specialized pre-training objective, which is specifically designed for summarization, i.e., generating salient sentences masked within the input document. The other two PLMs -- T5-base and BART-base -- also exhibit comparable results. Although LLMs are known for their strong performance across a wide range of NLP tasks, we clearly observe that smaller models fine-tuned on task-specific datasets can outperform them on the target task.

\begin{table*}[!h]
\centering
\caption{\small Evaluation of the quality of the generated titles for \textbf{LREC-COLING-2024}. Case 1 corresponds to fine-tuning on \textbf{CSPubSum} training set while Case 2 denotes fine-tuning on \textbf{SpringerSSAT} training set. All scores are reported as F1-scores in \%.}
\label{Table:par_all_types_rouge_meteor_bert-coling}
\begin{adjustbox}{width=1.0\linewidth}
{\begin{tabular}{|ccccccccc|} \hline
Model Name &Case &ROUGE-1 &ROUGE-2 &ROUGE-L &METEOR   &MoverScore &BERTScore &SciBERTScore \\\hline
T5-base &1 &46.84  &28.7 &41.69 &39.88 &39.6  &88.71 &76.05\\ \hline
T5-base &2 &44.76  &25.8 &38.01 &38.72 &36.64  &88.22 &73.69\\ \hline
BART-base &1 &46.87 &27.66 &41.89 &38.93 &39.61  &88.84 &75.94\\ \hline
BART-base &2 &42.43  &23.59 &36.23 &36.93 &35.32  &88.03 &72.85\\ \hline
PEGASUS-large &1 &\bf{49.85} &\bf{30.51} &\bf{43.93} &\bf{43.23} &\bf{41.66}  &89.1 &\bf{76.74}\\ \hline
PEGASUS-large &2 &47.64  &27.74 &40.56 &42.32 &38.98  &88.66 &74.79\\ \hline
LLaMA-3-8B$^\ast$ & &32.92 &16.66 &27.66 &30.61 &25.68  &86.77 &67.42 \\ \hline
LLaMA-3-8B &1 &45.3 &26.53 &40.51 &38.18 &38.93 &88.83 &76.14\\ \hline
LLaMA-3-8B  &2 &47.13  &27.23 &41.1 &40.38 &40.57  &89.5 &76.4\\ \hline
GPT-3.5-turbo & &45.16 &23.97 &38.88 &37.45 &38.85 &\bf{89.54} &75.64\\ \hline
\end{tabular} }
\end{adjustbox}
\end{table*}

\begin{table*}[!h]
\centering
\caption{\small Evaluation of factual consistency of the generated titles in \textbf{LREC-COLING-2024}. Case 1 corresponds to fine-tuning on \textbf{CSPubSum} training set while Case 2 denotes fine-tuning on \textbf{SpringerSSAT} training set. All scores are in \%.}
\label{Table:entity-scorecoling}
    \begin{adjustbox}{width=\linewidth}
{\tiny
{\begin{tabular}{|cccccccccc|} \hline

Model Name &Case &$prec_s^{NU}$ & $prec_s^{U}$& $prec_t^{NU}$& $recall_t^{NU}$ & $F1_t^{NU}$ & $prec_t^{U}$& $recall_t^{U}$& $F1_t^{U}$ \\ \hline
T5-base &1 &97.44 &97.39&63.06 &59.24 &57.17 &62.8 &58.88 &56.87 \\ \hline
T5-base &2 &97.41 &97.34 &60.34 &55.44 &54.73 &60.12 &54.95 &54.44\\ \hline
BART-base &1 &96.24 &96.23 &64.3 &59.21 &57.78 &64.11 &58.92 &57.56  \\ \hline
BART-base &2 &95.2 &95.17 &57.76 &54.21 &52.98 &57.41 &53.72 &52.56\\ \hline
PEGASUS-large &1 &\bf{97.87} &\bf{97.85} &\bf{66.32} &\bf{64.64} &\bf{61.47} &\bf{66.15} &\bf{64.30} &\bf{61.28} \\ \hline
PEGASUS-large &2 &97.41 &97.32 &60.46 &60.03 &57.13 &60.13 &59.67 &56.82\\ \hline
LLaMA-3-8B$^\ast$ & &74.6 &74.52 &45.29 &48.16 &43.48 &45.02 &47.64 &43.13 \\ \hline
LLaMA-3-8B &1 &81.57 &81.55 &62.18 &55.99 &55.19 &62.13 &55.95 &55.16 \\ \hline
LLaMA-3-8B &2 &88.25 &88.23 &64.17 &55.09 &56.14 &63.95 &54.93 &55.97\\ \hline
GPT-3.5-turbo & &91.98 &91.98 &60.87 &57.66 &55.56 &60.84 &57.69 &55.56 \\ \hline
\end{tabular} 
}
}
\end{adjustbox}
\end{table*}

Now, let us take a closer look at the performance tables, starting with the $n$-gram and semantic similarity between the author-written titles and the generated titles, as presented in Tables \ref{Table:par_all_types_rouge_meteor_bert-cs} and \ref{Table:par_all_types_rouge_meteor_bert-SpringerSSAT}. 
It is evident that {LLaMA-3-8B$^\ast$} (without fine-tuning) performs considerably worse than the other models, which have been fine-tuned on domain-specific data. This suggests that fine-tuning remains important for achieving good results, despite the extensive pre-training these models undergo. 

Interestingly, GPT-3.5-turbo performs better than LLaMA-3-8B$^\ast$, even in a zero-shot setting. One possible explanation is that GPT-3.5-turbo benefits from continuous worldwide usage, which effectively provides ongoing training. 

Another observation is that although fine-tuned {LLaMA-3-8B} and GPT-3.5-turbo achieve lower ROUGE and METEOR scores compared to some smaller pre-trained models, their performance according to semantic metrics such as BERTScore and SciBERTScore is closer to that of the latter. For example, in Table \ref{Table:par_all_types_rouge_meteor_bert-cs}, SciBERTScore values for all models except {LLaMA-3-8B$^\ast$} lie between 76.06 and 76.93. Similarly, in Table \ref{Table:par_all_types_rouge_meteor_bert-SpringerSSAT}, SciBERTScore ranges from 71.19 to 72.43 for all models except {LLaMA-3-8B$^\ast$}. Comparable trends can be observed for the MoverScore metric.

Overall, these results indicate that, except for {LLaMA-3-8B$^\ast$}, the models produce outputs that are semantically close to the original titles, even when word-level overlap is low. This highlights the highly abstractive nature of the LLM outputs. Similar patterns can also be seen in Table \ref{Table:par_all_types_rouge_meteor_bert-coling}.

Now, let us analyze the entity overlap between the generated titles and the author-written titles and abstracts. Table \ref{Table:entity-scoreCSPuBSum} displays the figures for {CSPubSum}. Similarly, Table \ref {Table:entity-SpringerSSAT} displays the results for the SpringerSSAT dataset. First note that the precision-source as well as F1-target do not vary much whether or not we count multiple occurrences of an entity. Again, {LLaMA-3-8B$^\ast$} (without fine-tuning) achieves the lowest scores for these metrics on both datasets. The smaller pre-trained models display similar performance to one another. Consider precision source $prec_s^{NU}$: its large value indicates that the entities in the generated summary are mostly present in the input source which is the abstract here. Table \ref{Table:entity-scoreCSPuBSum} shows that $prec_s^{NU} = 97.1$ for {T5-base}, $97.44$ for {BART-base}, and $98.13$ for {PEGASUS-large}. But $prec_s^{NU}$ is 92.73 for GPT-3.5-turbo and 78.95 for LLaMA-3-8B, showing that these models generate novel words more frequently. In contrast, the F1-target ($F1_t^{NU|U}$) lies in the 50's for all models, meaning that there is only moderate overlap between the entities in the ground-truth title and the generated title -- thus, a generated titles does not match the golden title very well but, at least for smaller models, it generally does not include entities outside the given abstract. Note that F1-target for LLMs is lower than that for the smaller models, again pointing to the abstractive nature of their output.   Clearly, the lower values of $prec_s^{NU|U}$ and $F1_t^{NU|U}$ for LLMs could indicate hallucination. But upon manually examining a few generated titles we did not notice any instances of hallucination. This hints at the presence of novel words that preserve the intended meaning.

\section{Case Studies}
\label{sec:caseStudies}
We now present examples of titles generated by the models from abstracts in the test sets of each of the three selected datasets.  

\subsection{Case Study from CSPubSum}
Table~\ref{fig:sample_Abs_title-cspubsum3} presents an example from CSPubSum test set. The titles generated by the fine-tuned T5-base, BART-base, and PEGASUS-large models exhibit significant similarity to the corresponding author-written references. {T5-base} has produced a very long title. Notably, {BART-base} and {PEGASUS-large} generated the same titles. 
In contrast, {LLaMA-3-8B$^*$} (without fine-tuning), generated a title that resembled a one-sentence summary and even that is incomplete; it appears more like an extract from the abstract. After fine-tuning, the {LLaMA-3-8B} model generated an acceptable title. GPT-3.5-turbo, using prompt-based techniques, successfully generated a title that captures the essence of the paper but contains novel words. 
Recollect that according to the evaluation metrics in Table \ref{Table:par_all_types_rouge_meteor_bert-cs}, the titles generated by {GPT-3.5-turbo} score lower than those generated by the fine-tuned {PEGASUS-large} model in terms of lexical overlap due to the presence of novel words. This discrepancy highlights a limitation of the automated metrics in accurately evaluating highly abstractive and stylistically rich titles.
\begin{table*}[!t]
 \centering 
 \caption{\small Input is an abstract from \textbf{CSPubSum} test set. Titles generated by the different models are shown. Fine-tuning was done on CSPubSum training set. Paper  taken from \texttt{\small \url{ https://www.sciencedirect.com/science/article/abs/pii/S037722171500586X}}.}	
 \label{fig:sample_Abs_title-cspubsum3}
 \begin{tabular}{|p{15.5 cm}|} \hline
 {\bf Author-written title:} {``Comparative statics effects independent of the utility function. When do we act the same way under risk?''} 
\\\hline
 {\bf T5-base:} ``Comparative statics effects independent of the utility function for portfolio choice and competitive firm under price uncertainty''\\\hline 	    
 {\bf BART-base:} ``Comparative statics effects in the context of expected utility''\\\hline
 {\bf PEGASUS-large:} ``Comparative statics effects in the context of expected utility''\\\hline
 {\bf  LLaMA-3-8B$^*$:} ``The author proposes a methodological approach that enables comparative static analysis of various economic models, irrespective of the''\\\hline
 {\bf LLaMA-3-8B:} ``Comparative statics analysis: A new approach''\\\hline
 {\bf Zero-shot GPT-3.5-turbo:} ``Comparative statics effects on expected utility in decision-making''\\\hline
 \end{tabular} 	
  
\end{table*}
\subsection{Case Study from SpringerSSAT}

In this sub-section, we present titles generated from a sample abstract in the SpringerSSAT test subset. These outputs, depicted in Table \ref{fig:sample_Abs_title-SpringerSSAT}, were produced by models fine-tuned on the corresponding training set.
In the case of {T5-base}, the generated title closely aligned with the author-written title.   
In case of {BART-base}, the generated title added a term ``understanding'' that reflects an investigative focus. In case of {PEGASUS-large}, the generated title matched almost perfectly with the human-authored title. In contrast, LLaMA-3-8B$^*$ (without fine-tuning) generated a title resembling a multi-sentence summary, which was also incomplete.
In case of {LLaMA-3-8B}, the generated title added the names of specific hospitals. Note that the abbreviations HC and STT were present in the abstract,  and the model correctly expanded them to their full forms,  Hwedza Central Hospital and St Theresa Tshita Hospital, respectively. This indicates an accurate interpretation of the source text. 

\begin{table*}[!h] 
 \centering 
  \caption{\small Input is an abstract from \textbf{SpringerSSAT} test set. Titles generated by the different models are shown. Fine-tuning was done on SpringerSSAT training set. Paper  taken from \texttt{\small \url{https://link.springer.com/article/10.1007/s43545-023-00821-5}}.}	
 \label{fig:sample_Abs_title-SpringerSSAT} 
 \begin{tabular}{|p{15.5 cm}|} \hline
 {\bf Author-written title:} {``Environmental health risks associated with solid waste management at rural hospitals in Chirumanzu District, Zimbabwe''} 
\\\hline
 {\bf T5-base:} ``Environmental health risks associated with hospital solid waste management strategies utilised at rural hospitals in Zimbabwe''\\\hline 	    
 {\bf BART-base:} ``Understanding the health risks associated with hospital solid waste management at rural hospitals in Zimbabwe''\\\hline
 {\bf PEGASUS-large:} ``Environmental health risks associated with solid waste management at rural hospitals in Zimbabwe''\\\hline
 {\bf  LLaMA-3-8B$^*$:} ``The authors' aim is to understand the environmental health risks associated with solid waste management strategies utilized at rural hospitals in Zimbabwe. They employ descriptive cross-sectional designs using both qualitative and quantitative approaches to collect, analyze, and present their findings. Data are gathered through''\\\hline
 {\bf LLaMA-3-8B:} ``Environmental health risk assessment of solid waste management practices at rural hospitals: case study of Hwedza Central Hospital and St Theresa Tshita Hospital''\\\hline
 \end{tabular} 	

\end{table*}

\subsection{Case Study on Cross-Domain Generalization to LREC-COLING-2024}

In this sub-section, we present representative examples of title generation from the LREC-COLING-2024 dataset in two different ways to illustrate the behavior of models fine-tuned on other datasets, namely CSPubSum and SpringerSSAT.

Table \ref{fig:sample_Abs_title-coling-1} shows the outputs generated by models for the LREC-COLING-2024 abstracts after fine-tuning on either  CSPubSum or SpringerSSAT. The model names and the labels (Case 1 or Case 2) indicate the corresponding setup:\\ 
\textbf{Case 1:} Models fine-tuned only on the CSPubSum training set.  \\
\textbf{Case 2:} Models fine-tuned only on the SpringerSSAT training set.\\

In Table \ref{fig:sample_Abs_title-coling-1}, for BART-base, titles are truncated in both cases due to the output token count limit. T5-base produces concise titles in both cases, but only Case 2 captures the \textit{low-resource} aspect present in the author-written title. {PEGASUS-large} includes `low-resource' accurately in both cases, with Case 1 better; in Case 2, the title is incomplete. The issue of truncated titles can be mitigated by increasing the token limit during inference. {LLaMA-3-8B} generates stable, readable, and identical titles across both cases but omits the low-resource detail. GPT-3.5-turbo produces a title similar to PEGASUS-large, using the abbreviation NMT, which may be less desirable for formal titles.

\begin{table*}[!h] 
\centering 
\caption{\small Input is an abstract from \textbf{LREC-COLING-2024}. Case 1 corresponds to fine-tuning on \textbf{CSPubSum} training set while Case 2 denotes fine-tuning on \textbf{SpringerSSAT} training set. Titles generated by the different models are shown. Paper taken from   \texttt{\url{https://aclanthology.org/2024.lrec-main.132/}}.}	
\label{fig:sample_Abs_title-coling-1} 
\begin{tabular}{|p{15.5 cm}|} \hline
{\bf Author-written title:} {``A Reinforcement Learning Approach to Improve Low-Resource Machine Translation Leveraging Domain Monolingual Data''} 
\\\hline
{\bf T5-base Case 1:} ``Reinforcement Learning Domain Adaptation for Neural Machine Translation''\\\hline 	  
{\bf T5-base Case 2:} ``Reinforcement Learning Domain Adaptation for Neural Machine Translation in the Low-Resource''\\\hline 
{\bf BART-base Case 1:} ``A novel Reinforcement Learning Domain Adaptation method for Neural Machine Translation in the low-''\\\hline
{\bf BART-base Case 2:} ``A novel Reinforcement Learning Domain Adaptation method for Neural Machine Translation (RLDA-''\\\hline
{\bf PEGASUS-large Case 1:} ``Reinforcement learning domain adaptation for Neural Machine Translation in the low-resource domain''\\\hline
{\bf PEGASUS-large Case 2:} ``Reinforcement learning domain adaptation for neural machine translation in low-resource domains: a novel''\\\hline
{\bf  LLaMA-3-8B$^*$:} ``Reinforced Domain Adaptation Method for Low Resource Neural Machine Translation''\\\hline
{\bf LLaMA-3-8B Case 1:} ``Reinforcement learning domain adaptation for neural machine translation''\\\hline
{\bf LLaMA-3-8B Case 2:} ``Reinforcement learning domain adaptation for neural machine translation''\\\hline
{\bf Zero-shot GPT-3.5-turbo:} ``Reinforcement Learning Domain Adaptation for Low-Resource NMT''\\\hline
\end{tabular} 	
\end{table*}

Table \ref{fig:sample_Abs_title-coling-2} shows the outputs of models fine-tuned on CSPubSum (Case 1) or SpringerSSAT (Case 2) for an abstract from the LREC-COLING-2024 dataset.
{T5-base} captures the ``entity abstraction approach'' in Case 1, but this highlights only one aspect of the model; Case 2 better reflects the overall methodology, though the title is truncated. {BART-base} produces an incomplete title in Case 1, omitting “for entailment tree generation,” while Case 2 improves coverage and explicitly mentions the task. PEGASUS-large and fine-tuned LLaMA-3-8B produce titles identical to the author-written title in both cases, demonstrating strong fidelity. LLaMA-3-8B without fine-tuning struggles to generate a concise title, whereas GPT-3.5-turbo emphasizes AI explanations, capturing the purpose of the model rather than the exact task.

\begin{table*}[!h]
 \centering 
  \caption{\small Input is an abstract from \textbf{LREC-COLING-2024}. Case 1 corresponds to fine-tuning on \textbf{CSPubSum} training set while Case 2 denotes fine-tuning on \textbf{SpringerSSAT} training set. Titles generated by the different models are shown.  Paper  taken from \texttt{\small \url{ https://aclanthology.org/2024.lrec-main.68/}}.}	
 \label{fig:sample_Abs_title-coling-2} 
 \begin{tabular}{|p{15.5 cm}|} \hline
 {\bf Author-written title:} {``A Logical Pattern Memory Pre-trained Model for Entailment Tree Generation''} 
\\\hline
{\bf T5-base Case 1:} ``An entity abstraction approach for logical pattern memory pre-trained models''\\\hline 	 
{\bf T5-base Case 2:} ``The logical pattern memory pre-trained model for entailment trees: a''\\\hline 
 {\bf BART-base Case 1:} ``Logical pattern memory pre-trained model''\\\hline
 {\bf BART-base Case-2:} ``A pre-trained model for generating logical entailment trees''\\\hline 
 {\bf PEGASUS-large Case-1:} ``A logical pattern memory pre-trained model for entailment tree generation''\\\hline
  {\bf PEGASUS-large Case-2:} ``A logical pattern memory pre-trained model for entailment tree generation''\\\hline
 {\bf  LLaMA-3-8B$^*$:} ``The proposed method addresses the limitations of previous approaches
by incorporating an external memory structure to capture the latent representations''\\\hline
 {\bf LLaMA-3-8B Case 1:} ``Logical pattern memory pre-trained model for entailment tree generation''\\\hline
  {\bf LLaMA-3-8B Case 2:} ``Logical Pattern Memory Pre-Trained Model: A New Approach for Generating Coherent Explainable Reasoning''\\\hline
 {\bf  Zero-shot GPT-3.5-turbo:} ``Improving AI Explanations with Logical Pattern Memory Pre-trained Model''\\\hline
 \end{tabular} 	

\end{table*}

\subsection{Manual Evaluation}
\label{sec:manualEval}
A human evaluation was conducted on thirty papers, comprising ten samples selected from each of the test subsets of CSPubSum and SpringerSSAT, and the LREC-COLING-2024 dataset. An annotator with expertise in NLP, currently pursuing a Master’s degree in Information Technology, was assigned to select the most appropriate title among those generated by the six models. For CSPubSum, PEGASUS-large was chosen as the preferred title in 80\% of the cases. In the SpringerSSAT dataset, PEGASUS-large remained the most preferred in 50\% of examples, followed by the fine-tuned LLaMA-3-8B in 30\% and T5-base in 20\% of cases. These results highlight PEGASUS-large’s consistent ability to generate coherent and contextually accurate titles. For the LREC-COLING-2024 dataset, PEGASUS-large was the winning model in 50\% of the cases, while GPT-3.5-turbo titles were selected as best in 40\% of the examples.

\section{Creative Titles with LLMs}
\label{sec:CreativeTitles}
We now explore whether ChatGPT can generate creative titles for research papers. It is challenging to train the smaller PLMs for this task due to the lack of suitable training data for this purpose. In contrast, LLMs, trained on massive and diverse corpora, possess the linguistic knowledge required to produce engaging and novel titles without the need for task-specific fine-tuning.

Specifically, we prompted \textbf{ChatGPT-4o} to generate titles in four distinct styles—\texttt{catchy}, \texttt{rhyming}, \texttt{funny}, and \texttt{comprehensive}—one style at a time for ten papers from the CSPubSum test set. For the SpringerSSAT test set, we employed \textbf{ChatGPT-5}. For LREC-COLING-2024, we  again employed \textbf{ChatGPT-4o}.

Table~\ref{fig:sample_Abs_title-CSPubSum-Style} presents an example from the CSPubSum dataset, displaying the four stylistic titles along with the `plain' title generated by GPT-3.5-turbo (following the method described in Section~\ref{sec:Impl} for CSPubSum dataset) and the author-written title. Table~\ref{fig:sample_Abs_title-SpringerSSAT-Style} shows a similar example for the SpringerSSAT dataset, including the `plain' title generated by ChatGPT-5  and the author-written title. Finally, Table~\ref{fig:sample_Abs_title-LREC-COLING-Style} presents the four title styles generated for an example from the LREC-COLING-2024 dataset, together with the `plain' title produced by GPT-3.5-turbo and the author-provided title.  

Table~\ref{Table:creative_auto_eval} summarizes the similarity between the generated titles and the author-written titles (ignoring the last column for now). Across CSPubSum, SpringerSSAT, and LREC-COLING-2024, \texttt{catchy}, \texttt{comprehensive}, and `plain' titles exhibit comparable and substantial lexical and semantic overlap with the author-written titles. In contrast, \texttt{rhyming} and \texttt{funny} titles generally share few words (sometimes with zero bigram overlap) and only moderate semantic similarity. This outcome is expected, as these stylistic formats often introduce novel words not present in the original titles and tend to omit technical terms to adhere to the chosen style.

\begin{table*}[!h]
\centering 
\caption{\small Titles generated by ChatGPT for an abstract in \textbf{CSPubSum}. Paper taken from  \texttt{\url{https://www.sciencedirect.com/science/article/abs/pii/S0010448513000535}}.}	
\label{fig:sample_Abs_title-CSPubSum-Style} 
\begin{tabular}{|p{15.5 cm}|} \hline
{\bf Author-written title:} {``Generic face adjacency graph for automatic common design structure discovery in assembly models''} 
\\\hline
{\bf Catchy:} ``Quantitative Assembly Model Reuse with GFAG for Common Design Structure Discovery''\\\hline 	    
{\bf Rhyming:} ``Graphs and Faces: Mapping Assemblies for Design Traces''\\\hline
{\bf Funny:} ``Graphing the Love: Finding Mating Faces in Assembly Design!''\\\hline
{\bf Comprehensive:} ``Generic Face Adjacency Graph for Common Design Structure Discovery in Assemblies''\\\hline
{\bf Plain:} ``Generic Face Adjacency Graph for Assembly Model Structure Discovery''\\\hline
\end{tabular} 	

\end{table*}

\begin{table*}[!h]
\centering 
\caption{\small Titles generated by ChatGPT for an abstract in \textbf{SpringerSSAT}. Paper taken from  \texttt{\url{https://link.springer.com/article/10.1007/s43545-021-00194-7}}.}	
\label{fig:sample_Abs_title-SpringerSSAT-Style} 
\begin{tabular}{|p{15.5 cm}|} \hline
{\bf Author-written title:} {``How do teachers perceptions of climate change vary in terms of importance, causes, impacts and mitigation? A comparative study in Bangladesh''} 
\\\hline
{\bf Catchy:} ``Teachers’ Perceptions of Climate Change in Sylhet: Implications for Education and Mitigation Strategies''\\\hline 	    
{\bf Rhyming:} ``Teaching the Change, Perceptions Arrange: Climate Awareness Among Sylhet’s Educators''\\\hline
{\bf Funny:} ``Teaching the Heat: How Sylhet’s Teachers See Climate Change''\\\hline
{\bf Comprehensive:} ``Teachers’ Perceptions of Climate Change in Bangladesh: Institutional, Sociodemographic, and Educational Influences''\\\hline
{\bf Plain:} ``Teachers’ Perceptions of Climate Change in Public and Private Schools in Sylhet, Bangladesh''\\\hline
\end{tabular} 	
\end{table*}

\begin{table*}[!h] 
\centering 
\caption{\small Titles generated by ChatGPT for an abstract in \textbf{LREC-COLING-2024}. Paper taken from  \texttt{\url{https://aclanthology.org/2024.lrec-main.150/}}.}	
\label{fig:sample_Abs_title-LREC-COLING-Style} 
\begin{tabular}{|p{15.5 cm}|} \hline
{\bf Author-written title:} {``A Tool for Determining Distances and Overlaps between Multimodal Annotations''} 
\\\hline
{\bf Catchy:} ``Analyzing ELAN Annotations: A Tool for Comparing Multimodal Overlaps in Corpus Analysis''\\\hline 	    
{\bf Rhyming:} ``Comparative Insights in ELAN's Light: A Tool for Annotations That Delivers Clear Sight''\\\hline
{\bf Funny:} ``Annotation Adventures: Finding Overlaps and Other Mischief!''\\\hline
{\bf Comprehensive:} ``Tool for Analyzing ELAN Annotations: Comparing Speech and Gesture Overlap in Multimodal Corpora''\\\hline
{\bf Plain:} ``Tool for Comparing ELAN Annotations in Multimodal Corpus Analysis''\\\hline
\end{tabular} 	
\end{table*}

\begin{table*}[!h]
\centering
\caption{\small Evaluation of ChatGPT-generated creative titles. All scores are reported as F1-scores in \%.} 
\label{Table:creative_auto_eval}
\begin{adjustbox}{width=1.0\linewidth}
\begin{tabular}{|c|cccccccc|c|} \hline
Dataset~ & Style &ROUGE-1 &ROUGE-2 &ROUGE-L &METEOR   &MoverScore &BERTScore &SciBERTScore & Avg. Rating\\\hline

\textbf{CSPubSum}~ & Catchy&45.81 &18.44 &38.83 &45.55 &35.43 &88.00  &75.77 & 4\\ 
         & Rhyming&13.20 &0 &11.37 &10.37 &12.32 &83.47 &61.02 & 3.3\\ 
         & Funny&18.57 &28.18 &14.95 &15.09 &14.76 &83.78 &59.53 & 2.8\\ 
         & Compreh.& \textbf{48.03} & \textbf{29.32} & 40.88 &\textbf{47.93} &\textbf{44.21} &\textbf{88.16} &76.78 & \textbf{4.4}\\ 
         & Plain&47.51 &25.03 &\textbf{42.09} &44.99 &40.74 &88.15 &\textbf{77.00} & 3.4\\ \hline\hline

\textbf{SpringerSSAT}~ & Catchy&40.6 &14.6 &31.63 &32.85 &28.14 &88.18 &70.62 &3.9 \\ 
                 & Rhyming &28.33 &2.65 &22.7 &20.33 &18.06 &86.78  &65.73 &2.8 \\ 
                 & Funny &25.72 &5.95 &22.31 &20.85 &14.75 &86.16  &64.16 &2.3\\ 
                 & Compreh.&42.67 &18.66 &33.62 &\bf{39.19} &28.98 &88.25  &71.23 &\bf{4.5}\\ 
                 & Plain &\bf{47.29} &\bf{21.54} &\bf{39.71} &35.86 &\bf{34.68} &\bf{89.13}  &\bf{73.55} &3.7\\ \hline

\textbf{LREC-COLING-2024}~ & Catchy&47.82 &\textbf{24.47} &\textbf{41.60} &45.61 &42.84 &90.07  &75.77 & 3.8\\ 
                 & Rhyming &20.48 &2.65 &16.94 &16.64 &20.72 &84.37  &61.02 & 3.1\\ 
                 & Funny&15.71 &0.0 &11.36 &10.11 &11.47 &84.05  &59.53 & 3.6\\ 
                 & Compreh.&\textbf{48.22} &21.88 &41.12 &\textbf{48.83} &\textbf{45.71} &90.40  &76.78 & \textbf{4.1}\\ 
                 & Plain&43.83 &19.45 &39.76 &39.42 &42.48 &\textbf{90.88}  &\textbf{77.00} & 3.7\\ \hline
\end{tabular} 
\end{adjustbox}
\end{table*}

We selected nine human annotators, all undergraduate or graduate students in computer science, to rate the ChatGPT-generated and author-written titles on a scale of 1–5. Each abstract and its corresponding generated and author-written titles (without revealing which one is author-written) were provided to three annotators randomly selected from the group.  The average ratings for ChatGPT-generated titles are reported in the last column of Table~\ref{Table:creative_auto_eval}. Among the four styles, the comprehensive titles received the highest scores. Author-written titles for all three datasets—LREC-COLING-2024, CSPubSum, and SpringerSSAT—obtained an average rating of 4. When asked to choose the best title for each paper, the annotators preferred ChatGPT-generated titles over the author-written ones for nine LREC-COLING-2024 papers, seven CSPubSum papers, and five SpringerSSAT papers, with a clear preference for the comprehensive or catchy styles.


\section{Expert-Written Alternative Titles}~\label{ExpertEvaluation}

A single ``correct'' title rarely exists for any given paper. Although we assumed the author-written titles to be the gold standard for our quantitative analysis, this reference represents only one of many possible ways to model the problem. Different authors would likely title the same abstract differently. To better capture this linguistic variability and address RQ5, we collected a set of alternative titles written by human experts. In particular, we randomly selected 10 samples from the test set of each dataset. For each of these three newly constructed subsets, three participants (hereafter referred to as \textit{experts}) were asked to independently read the abstracts and propose original titles. Note that they were not shown the author-written titles. Each participant was either a current Master’s student or a recent Master’s graduate, with a specialization in computer science and NLP. In total, nine experts annotated the selected abstracts with titles.
We then computed similarity scores using ROUGE, METEOR, and BERTScore along three dimensions: (i) Author vs Model, (ii) Expert vs Model, and (iii) Human vs Human, which includes two sub-cases: Author vs Expert, and Expert vs Expert. Note that we selected only fine-tuned PEGASUS-large and LLaMA-3-8B as the models for this experiment.

Tables \ref{Table:exprt_eval_CSPubSum}, \ref{Table:exprt_eval_SpringerSSAT} and \ref{Table:exprt_eval_LREC-COLING-2024} show that, for both models, similarity with the Author is generally higher than similarity with the other human Experts. This is expected as the models are trained on author-written titles. Human–human agreement is lower than Model–Author agreement, further reflecting the natural variability in human-written titles. Despite differences in $n$-gram metrics (ROUGE), BERTScore remains high across all comparisons, suggesting that both models and humans capture the underlying meaning of the abstract even when their wording differs.

\begin{table*}[!h]
\centering
\caption{\small Model vs human, and human vs human evaluation on 10 selected examples from \textbf{CSPubSum} test set. The models are fine-tuned on \textbf{CSPubSum} training set. All scores are reported as F1-scores in \%.}
\label{Table:exprt_eval_CSPubSum}
\begin{adjustbox}{width=\linewidth}
\begin{tabular}{|lccccc|} \hline
\textbf{Comparison} & \textbf{ROUGE-1} & \textbf{ROUGE-2} & \textbf{ROUGE-L} & \textbf{METEOR} & \textbf{BERTScore} \\\hline
Author vs PEGASUS-large  &51.18  &34.11  &50.38  &49.14 &90.85  \\
Expert-1 vs PEGASUS-large &41.17  &22.49  &38.11  &37.11 &90.42  \\
Expert-2 vs PEGASUS-large  &36.59  &13.8  &32.77  &29.11 &88.56  \\
Expert-3 vs PEGASUS-large &48.1  &29.11  &43.99  &38.08 &90.25  \\\hline

Author vs LLaMA-3-8B &46.23  &30.07  &45.43  &43.71 &90.11  \\
Expert-1 vs LLaMA-3-8B &41.15  &19.51  &38.31  &36.34 &91.0  \\
Expert-2 vs LLaMA-3-8B &25.38  &5.24  &22.27  &18.11 &87.74  \\
Expert-3 vs LLaMA-3-8B &36.67  &18.97  &34.22  &29.6 &89.35  \\\hline

Author vs Expert-1 &39.32  &20.75  &36.04  &39.76 &89.14  \\
Author vs Expert-2 &27.18  &10.18  &24.32  &27.98 &88.3  \\
Author vs Expert-3 &41.48  &20.71  &37.81  &45.9 &89.14  \\\hline

Expert-1 vs Expert-2 &34.97  &14.39  &31.46  &26.73 &87.71  \\
Expert-1 vs Expert-3 &39.0  &15.69  &34.34  &38.41 &89.15  \\
Expert-2 vs Expert-3 &39.51  &13.73  &31.36  &37.84 &88.56  \\\hline
\end{tabular} 
\end{adjustbox}
\end{table*}

\begin{table*}[!ht]
\centering
\caption{\small Model vs human, and human vs human evaluation on 10 selected examples from \textbf{SpringerSSAT} test set. The models are fine-tuned on \textbf{SpringerSSAT} training set. All scores are reported as F1-scores in \%.}
\label{Table:exprt_eval_SpringerSSAT}
\begin{adjustbox}{width=1.0\linewidth}
\begin{tabular}{|lccccc|} \hline
\textbf{Comparison} & \textbf{ROUGE-1} & \textbf{ROUGE-2} & \textbf{ROUGE-L} & \textbf{METEOR} & \textbf{BERTScore} \\\hline
Author vs PEGASUS-large &51.18  &34.11  &50.38  &49.14 &90.85  \\
Expert-1 vs PEGASUS-large &41.17  &22.49  &38.11  &37.11 &90.42  \\
Expert-2 vs PEGASUS-large &36.59  &13.8  &32.77  &29.11 &88.56  \\
Expert-3 vs PEGASUS-large &48.1  &29.11  &43.99  &38.08 &90.25  \\\hline

Author vs LLaMA-3-8B &46.23  &30.07  &45.43  &43.71 &90.11  \\
Expert-1 vs LLaMA-3-8B  &41.15  &19.51  &38.31  &36.34 &91.0  \\
Expert-2 vs LLaMA-3-8B &25.38  &5.24  &22.27  &18.11 &87.74  \\
Expert-3 vs LLaMA-3-8B  &36.67  &18.97  &34.22  &29.67 &89.35  \\\hline

Author vs Expert-1 &39.32  &20.75  &36.04  &39.76 &89.14  \\
Author vs Expert-2  &27.18  &10.18  &24.32  &27.98 &88.3  \\
Author vs Expert-3 &41.48  &20.71  &37.81  &45.98  &89.14  \\\hline

Expert-1 vs Expert-2 &34.97  &14.39  &31.46  &26.73 &87.71  \\
Expert-1 vs Expert-3 &39.0  &15.69  &34.34  &38.41 &89.15  \\
Expert-2 vs Expert-3 &39.51  &13.73  &31.36  &37.84 &88.56  \\\hline
\end{tabular} 
\end{adjustbox}
\end{table*}

\begin{table*}[!b]
\centering
\caption{\small Model vs human, and human vs human evaluation on 10 selected examples from \textbf{LREC-COLING-2024} dataset. The models are fine-tuned on \textbf{CSPubSum} training set. All scores are reported as F1-scores in \%.} 
\label{Table:exprt_eval_LREC-COLING-2024}
\begin{adjustbox}{width=1.0\linewidth}
\begin{tabular}{|lccccc|} \hline
\textbf{Comparison} & \textbf{ROUGE-1} & \textbf{ROUGE-2} & \textbf{ROUGE-L} & \textbf{METEOR} & \textbf{BERTScore} \\\hline
Author vs PEGASUS-large & 51.18 & 34.11 & 50.38 & 49.14 & 90.85 \\
Expert-1 vs PEGASUS-large & 4117 & 22.49 & 38.11 & 37.11 & 90.42  \\
Expert-2 vs PEGASUS-large & 36.59 & 13.8 & 32.77 & 29.11 & 88.56 \\
Expert-3 vs PEGASUS-large  & 48.1 & 29.11 & 43.99 & 38.08 & 90.25 \\\hline

Author vs LLaMA-3-8B & 46.23 & 30.07 & 45.43 & 43.71 & 90.11 \\
Expert-1 vs LLaMA-3-8B & 41.15 & 19.51 & 38.31 & 36.34 & 91.0 \\
Expert-2 vs LLaMA-3-8B & 25.38 & 5.24 & 22.27 & 18.11 & 87.74\\
Expert-3  vs LLaMA-3-8B & 36.67 & 18.97 & 34.22 & 29.67 & 89.35 \\\hline

Author vs Expert-1 & 39.32 & 20.75 & 36.04 & 39.76 & 89.14 \\
Author vs Expert-2 & 27.18 & 10.18 & 24.32 & 27.98 & 88.3 \\
Author vs Expert-3 & 41.48 & 20.71 & 37.81 & 45.98 & 89.14 \\\hline

Expert-1 vs Expert-2 & 34.97 & 14.39 & 31.46 & 26.73 & 87.71 \\
Expert-1 vs Expert-3 & 39.0 & 15.69 & 34.34 & 38.41 & 89.15 \\
Expert-2 vs Expert-3 & 39.51 & 13.73 & 31.36 & 37.84 & 88.56 \\\hline
\end{tabular} 
\end{adjustbox}
\end{table*}

Tables~\ref{fig:sample_ExpertEval-cspubsum1} and \ref{fig:sample_ExpertEval-cspubsum2} provide a qualitative comparison between author-written titles, titles generated by the fine-tuned models (PEGASUS-large and LLaMA-3-8B), and those proposed by human experts. The examples again illustrate that model-generated titles tend to show closer alignment with the author-written titles compared to expert written alternatives. The input abstracts are taken from the test set of CSPubSum. Tables~\ref{fig:sample_ExpertEval-cspubsum1} and \ref{fig:sample_ExpertEval-cspubsum2} show examples where the titles written by Expert-2 are considerably longer and more detailed than those from other sources. Although this reflects strong subject knowledge, the title becomes less concise and departs from the compact style typically observed in author-written ones. 
This difference also suggests that novice researchers may struggle to formulate clear and well-balanced titles. In this context, automated title generation can serve as a useful support tool.
Similar illustrations are provided in Tables~\ref{fig:sample_ExpertEval-SpringerSSAT1} and~\ref{fig:sample_ExpertEval-SpringerSSAT2} (SpringerSSAT) and Tables~\ref{fig:sample_ExpertEval-LREC-COLING-2024-1} and \ref{fig:sample_ExpertEval-LREC-COLING-2024-2} (LREC-COLING-2024).

\begin{table*}[!htbp]
 \centering 
 \caption{\small Comparison of author-written, model-generated (PEGASUS-large and LLaMA-3-8B fine-tuned on \textbf{CSPubSum} training set), and expert-written titles for an abstract from \textbf{CSPubSum} test set. Paper  taken from \texttt{\small \url{ https://www.sciencedirect.com/science/article/pii/S0010448513000675}}.}
 \label{fig:sample_ExpertEval-cspubsum1} 
 \begin{tabular}{|p{15.5 cm}|} \hline
 {\bf Author-written title:} ``Integrated construction and simulation of tool paths for milling dental crowns and bridges'' \\\hline 
 {\bf Fine-tuned PEGASUS-large:} ``Tool path generation and simulation for dental implant milling using an isoparametric and isogeodesic approach''\\\hline
 {\bf Fine-tuned LLaMA-3-8B:} ``Integrated tool path generation and simulation for milling in dental technology''\\\hline
 {\bf Expert-1:} ``An Novel approach in construction of tool path generation and simulation for milling in the field of dental technology''\\\hline
 {\bf Expert-2:} ``Optimization of tool path generation using isoparametric and isogeodesic approach by evaluating feed rate selection and distance error of the machine surface to increase robustness and universality for milling in the field of dental technology''\\\hline
 {\bf Expert-3:} ``Tool Path Generation and Simulation: A Novel Approach for Milling Process in Dental Technology''\\\hline
 \end{tabular} 	
\end{table*}

\begin{table*}[!htbp] 
 \centering 
 \caption{\small Comparison of author-written, model-generated (PEGASUS-large and LLaMA-3-8B fine-tuned on \textbf{CSPubSum} training set), and expert-written titles for an abstract from  CSPubSum test set.  Paper  taken from \texttt{\small \url{ https://www.sciencedirect.com/science/article/pii/S0010448513001565}}.}
 \label{fig:sample_ExpertEval-cspubsum2} 
 \begin{tabular}{|p{15.5 cm}|} \hline
 {\bf Author-written title:} ``A parallel algorithm for improving the maximal property of Poisson disk sampling'' \\\hline    
 {\bf Fine-tuned PEGASUS-large:} ``A three-stage solution process for the integrated design and control problem in paper mill design''\\\hline
 {\bf Fine-tuned LLaMA-3-8B:} ``A fast algorithm for generating maximal Poisson disk samplings ''\\\hline
 {\bf Expert-1:} ``Implementing three-stage solution process to understand and analysis the Optimization of complex design problems''\\\hline
 {\bf Expert-2:} ``A three-stage multiobjective optimization approach for solving computationally expensive simulation-based design problems with conflicting objectives and decision maker preferences''\\\hline
 {\bf Expert-3:} ``Three-Stage Solution Process addressing Computational Challenges involving Multiple Conflicting Objectives in Complex Simulation-based Design Optimization Problems''\\\hline
 \end{tabular} 	
 \end{table*}

\begin{table*}[!htbp]
 \centering 
 \caption{\small Comparison of author-written, model-generated (PEGASUS-large and LLaMA-3-8B fine-tuned on \textbf{SpringerSSAT} training set), and expert-written titles for an abstract from SpringerSSAT test set. Paper  taken from \texttt{\small \url{https://link.springer.com/article/10.1007/s43545-024-00886-w}}.}
 \label{fig:sample_ExpertEval-SpringerSSAT1} 
 \begin{tabular}{|p{15.5 cm}|} \hline
 {\bf Author-written title:} ``The Woke way or the highway: American democracy in the age of Wokism and Cancel Culture'' \\\hline    
 {\bf Fine-tuned PEGASUS-large:} ``Wokism and Cancel Culture in the Context of Democracy''\\\hline
 {\bf Fine-tuned LLaMA-3-8B:} ``Wokeism and cancel culture: Concept appropriation as a threat to democracy''\\\hline
 {\bf Expert-1:} ``Concept Appropriation and Racialization: How Wokism and Cancel Culture Reshape Understanding of Race and Democracy''\\\hline
 {\bf Expert-2:} ``The Evolution of Cancel Culture: Re-characterization of Racialization and It's impact on Democracy''\\\hline
 {\bf Expert-3:} ``Emergence of Racialization within Anti-democratic Current Social Activism Trends''\\\hline
 \end{tabular} 	
 \end{table*}

\begin{table*}[!htbp]
 \centering 
\caption{\small Comparison of author-written, model-generated (PEGASUS-large and LLaMA-3-8B fine-tuned on \textbf{SpringerSSAT} training set), and expert-written titles for an abstract from SpringerSSAT test set. Paper  taken from \texttt{\small \url{https://link.springer.com/article/10.1023/B:SORE.0000041293.24615.f7}}.}
 \label{fig:sample_ExpertEval-SpringerSSAT2} 
 \begin{tabular}{|p{15.5 cm}|} \hline
 {\bf Author-written title:} ``The Moral Superiority Effect: Self Versus Other Differences in Satisfaction with Being Overpaid'' \\\hline    
 {\bf Fine-tuned PEGASUS-large:} ``Moral superiority effects on people's reactions to being overpaid''\\\hline
 {\bf Fine-tuned LLaMA-3-8B:} ``Moral Superiority Effects: People Think Others Are More Satisfied With Being Overpaid Than They Are Themselves''\\\hline
 {\bf Expert-1:} ``The Moral Superiority Effect: Discrepancies in Perceived Satisfaction with Overpayment''\\\hline
 {\bf Expert-2:} ``The Moral Superiority Effect in Reactions to Overpayment''\\\hline
 {\bf Expert-3:} ``Biased Perception of Moral Superiority between Self and Others in Overpayment''\\\hline
 \end{tabular} 	
 
\end{table*}

\begin{table*}[!htbp]
 \centering 
 \caption{\small Comparison of author-written titles, model-generated titles (from PEGASUS-large and LLaMA-3-8B fine-tuned on the \textbf{CSPubSum} training set), and expert-written titles for an abstract from the LREC-COLING-2024 dataset. Paper taken from \url{https://aclanthology.org/2024.lrec-main.30/}.}
 \label{fig:sample_ExpertEval-LREC-COLING-2024-1}
 \begin{tabular}{|p{15.5 cm}|} \hline
 {\bf Author-written title:} ``Active Learning Design Choices for NER with Transformers'' \\\hline    
 {\bf Fine-tuned PEGASUS-large:} ``Evaluating sub-sentence annotation strategies for token classification using transformer networks''\\\hline
 {\bf Fine-tuned LLaMA-3-8B:} ``Active learning for named entity recognition using transformers''\\\hline
 {\bf Expert-1:} ``Handling partial annotations in active learning for transformer token classification''\\\hline
 {\bf Expert-2:} ``Optimizing Annotation Strategies through Efficient Transformer based Token Classification''\\\hline
 {\bf Expert-3:} ``A comparative analysis of various annotation strategies for active learning in token classification using transformer networks''\\\hline
 \end{tabular} 	
 \end{table*}

\begin{table*}[!htbp] 
 \centering 
\caption{\small Comparison of author-written titles, model-generated titles (from PEGASUS-large and LLaMA-3-8B fine-tuned on the \textbf{CSPubSum} training set), and expert-written titles for an abstract from the LREC-COLING-2024 dataset. Paper taken from \url{https://aclanthology.org/2024.lrec-main.77/}.}
 \label{fig:sample_ExpertEval-LREC-COLING-2024-2} 
 \begin{tabular}{|p{15.5 cm}|} \hline
 {\bf Author-written title:} ``A Multimodal In-Context Tuning Approach for E-Commerce Product Description Generation'' \\\hline    
 {\bf Fine-tuned PEGASUS-large:} ``Multimodal In-Context Tuning for product descriptions''\\\hline
 {\bf Fine-tuned LLaMA-3-8B:} ``Multimodal in-context tuning: Generating accurate and diverse product descriptions''\\\hline
 {\bf Expert-1:} ``Multimodal product description generation using in context learning with visual and keyword inputs''\\\hline
 {\bf Expert-2:} ``ModICT: An efficient approach for enhancing Image-based Product Description Generation''\\\hline
 {\bf Expert-3:} ``ModICT : A Multimodal In-Context Tuning approach for generating product descriptions from images''\\\hline
 \end{tabular} 	
 \end{table*}

\section{Limitations}~\label{Limitations}
Although we have conducted a comprehensive study of automatic title generation from scientific abstracts, this research is not without limitations. First, the datasets are limited to a few selected fields, primarily computer science (CSPubSum), natural language processing (LREC-COLING-2024), and social sciences (SpringerSSAT). Therefore, the results may not generalize to other disciplines where title conventions differ significantly, such as medicine, life sciences, physics, chemistry, or mathematics. Second, non-english publications and cultural variations are not captured, limiting the applicability of our findings in multilingual contexts. Third, we evaluated only a limited set of PLMs and LLMs, and their performance may differ significantly from other models reported in the literature. Moreover, these models evolve rapidly, and the performance of the newer versions may deviate significantly from the results presented in this paper.

\section{Conclusion}~\label{conclusion}
This study examined automatic research title generation using pre-trained language models (T5-base, BART-base, PEGASUS-large) and large language models (LLaMA-3-8B, GPT-3.5-turbo). A new domain-specific dataset, SpringerSSAT, has been introduced for fine-tuning and evaluation, while the LREC-COLING-2024 corpus was used only for cross-domain testing without additional training.

Results show that PEGASUS-large consistently achieved the best performance across automatic and human evaluation metrics while being more efficient than larger LLMs. Fine-tuning on domain-relevant data enhances the coherence, factual accuracy, and relevance of generated titles. The models also demonstrated strong generalization ability across datasets.

Human evaluation indicates that AI-generated titles are comparable to author-written ones in terms of creativity and clarity, though minor editing may further refine stylistic aspects. All developed models and datasets are publicly available to encourage further research.

Future work will focus on multilingual adaptation, retrieval-based generation for factual consistency, and improved evaluation methods for creativity and abstraction. Overall, the study demonstrates that well-tuned mid-sized models can generate accurate and meaningful academic titles with lower computational cost than very large language models.

\vspace{1mm}
\bibliographystyle{unsrt}
\bibliography{anthology}

\end{document}